\documentclass{article}
\usepackage{spconf,amsmath,graphicx}
\usepackage{pifont}
\newcommand{\cmark}{\ding{51}}
\newcommand{\xmark}{\ding{55}}
\usepackage{xcolor}
\usepackage{svg}
\usepackage{soul}
\usepackage{booktabs}
\usepackage{mathtools, nccmath}
\usepackage{amsfonts}

\usepackage[activate={true,nocompatibility},spacing=true,final,tracking=true,kerning=true,factor=1100,stretch=10,shrink=10]{microtype}

\usepackage{multirow}

\title{a simplified fully quantized transformer for end-to-end speech recognition}
\name{Alex Bie\sthanks{Work performed during an internship at Huawei Noah's Ark Lab.}
, Bharat Venkitesh, Joao Monteiro, Md Akmal Haidar, Mehdi Rezagholizadeh}
\address{Huawei Noah's Ark Lab, Montreal Research Centre, Canada \\
{\fontsize{10pt}{10pt}\selectfont alexbie98@gmail.com, \{bharat.venkitesh, joao.monteiro, md.akmal.haidar, mehdi.rezagholizadeh\}@huawei.com}}

\begin{document}
\maketitle

\ninept
\begin{abstract}
While significant improvements have been made in recent years in terms of end-to-end automatic speech recognition (ASR) performance, such improvements were obtained through the use of very large neural networks, unfit for embedded use on edge devices. That being said, in this paper, we work on simplifying and compressing Transformer-based encoder-decoder architectures for the end-to-end ASR task. We empirically introduce a more compact Speech-Transformer by investigating the impact of discarding particular modules on the performance of the model. Moreover, we evaluate reducing the numerical precision of our network's weights and activations while maintaining the performance of the full-precision model. Our experiments show that we can reduce the number of parameters of the full-precision model and then further compress the model 4x by fully quantizing to 8-bit fixed point precision. 

\end{abstract}
\vspace{-5pt}
\section{Introduction}
\label{sec:intro}


End-to-end automatic speech recognition (ASR) systems combine the functionality of acoustic, pronunciation, and language modelling components into a single neural network. Early approaches to end-to-end ASR employ CTC \cite{graves2006connectionist, graves2014towards}; however these models require rescoring with an external language model (LM) to obtain good performance \cite{bahdanau2016end}.
RNN encoder-decoder \cite{cho-etal-2014-learning,sutskever2014sequence} equipped with attention \cite{bahdanau2014neural}, originally proposed for machine translation, is an effective approach for end-to-end ASR \cite{bahdanau2016end, chan2016listen}. These systems see less of a performance drop in the no-LM setting \cite{bahdanau2016end}. 
 
 More recently, the Transformer \cite{vaswani2017attention} encoder-decoder architecture has been applied to ASR \cite{dong2018speech, mohamed2019transformers, karita2019comparative}. Transformer training is parallelizable across time, leading to faster training times than recurrent models~\cite{vaswani2017attention}. This makes them especially amenable to the large audio corpora encountered in speech recognition. Furthermore, Transformers are powerful autoregressive models \cite{radford2019language, parmar2018image}, and have achieved reasonable ASR results without incurring the storage and computational overhead associated with using LM's during inference \cite{mohamed2019transformers}.

Although current end-to-end technology has seen significant improvements in accuracy, computational requirements in terms of both time and space for performing inference with such models remains prohibitive for edge devices. Thus, there has been increased interest in reducing model sizes to enable on-device computation. The model compression literature explores many techniques to tackle the problem, including: quantization \cite{jacob2018quantization}, pruning \cite{lecun1990optimal, han2015deep}, and knowledge distillation \cite{kim2019knowledge, kim2016sequence}. An all-neural, end-to-end solution based on RNN-T ~\cite{graves2012sequence} is presented in \cite{He2018}. The authors make several runtime optimizations to inference and perform post-training quantization, allowing the model to be successfully deployed to edge devices.

\begin{figure}[t]
    \includegraphics[scale=0.5]{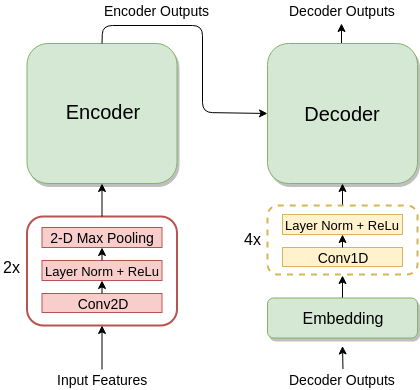}
    \vspace{-5pt}
    \caption{\ninept Components of end-to-end Transformer ASR as described in ~\cite{mohamed2019transformers}. 2D convolutional blocks are used for feature extraction and down-sampling. 1D causal convolutions are applied on the decoder side. Our proposed simplified Transformer only uses the \textbf{green} and \textbf{red} blocks (no decoder causal convolutions or sinusoidal positional encodings); as such, the decoder receives \textit{no explicit positional information}.}
    \label{fig:block_diagram}
    \vspace{-10pt}
\end{figure}

In this contribution, we turn our focus to refining the Transformer architecture so as to enable its use on edge devices. The absence of recurrent connections in Transformers provides a significant advantage in terms of speeding up computation, and therefore, quantizing a Transformer-based ASR system would be an important step towards on-device ASR.
We report findings on direct improvements to the model through removing components which do not significantly affect performance, and finally reduce the numerical precision of model weights and activations. Specifically, we reduced the dimensionality of the inner representations throughout the model, removed convolutional layers employed prior to the decoder's layers (as in Fig.~\ref{fig:block_diagram}), and finally performed 8-bit quantization to the model's weights and activations (following ~\cite{prato2019fully} which introduces a fully quantized transformer for machine translation and language modeling). As verified in terms of recognition performance, our results on the Librispeech dataset~\cite{panayotov2015librispeech} support the claim that one can recover the original performance even after greatly reducing model's computational requirements.


The remainder of this work is organized as follows: section~\ref{sec:transformer} gives an overview of Transformer-based ASR, and section~\ref{sec:quantization} describes the details of the quantization scheme. Section \ref{sec:experiments} describes our experiments with the Librispeech dataset. Section \ref{sec:discussion}  is a discussion of our results. Connection to prior work is presented in section~\ref{sec:prior}. Finally, we draw conclusions and describe future directions in section~\ref{sec:conclusion}.


\section{Transformer networks for ASR}
\label{sec:transformer}
Casting ASR as a sequence-to-sequence task, the Transformer encoder takes as input a sequence of frame-level acoustic features $(x_1,...,x_T)$, and maps it to a sequence of high-level representations $(h_1,...,h_N)$. The decoder generates a transcription $(y_1,...,y_L)$ one token at a time. Each choice of output token $y_l$ is conditioned on the hidden states $(h_1,...,h_N)$ and previously generated tokens $(y_1,...,y_{l-1})$ through attention mechanisms. The typical choice for acoustic features are frame-level log-Mel filterbank coefficents. The target transcripts are represented by word-level tokens or sub-word units such as characters or produced through byte pair encoding \cite{sennrich2016neural}.



\vspace{-5pt}
\subsection{Transformer architecture}
The encoder and decoder of the Transformer are stacks of $N$ Transformer layers.
The layers of the encoder iteratively refine the representation of the input sequence with a combination of multi-head self-attention and frame-level affine transformations. Specifically, the inputs to each layer are projected into keys $K$, queries $Q$, and values $V$. Scaled dot product attention is then used to compute a weighted sum of values for each query vector:
\begin{align} 
\operatorname{Attention}(Q,K,V) = \operatorname{softmax}(\frac{QK^T}{\sqrt{d_k}})V
\end{align}
where $d_k$ is the dimension of the keys. We obtain multi-head attention by performing this computation $h$ times independently with different sets of projections, and concatenating:
\begin{align} 
\operatorname{MultiHead}(Q,K,V) = \operatorname{Concat}(\operatorname{head_1},...,\operatorname{head_h})W^O \\
\operatorname{head_i} = \operatorname{Attention}(QW_i^Q,KW_i^K,VW_i^V)
\end{align}
The $W_i^*$ are learned linear transformations $W_i^*:d_{model} \rightarrow d_*$, and $W^O : h \cdot d_v \rightarrow d_{model}$. We use $d_* = d_{model}/h$. The self-attention operation allows frames to gather context from all timesteps and build an informative sequence of high-level features. The outputs of multi-head attention  go through a 2-layer position-wise feed-forward network with hidden size $d_{ff}$.
\begin{align} 
\operatorname{FFN}(x) = W_2\operatorname{ReLu}(W_1x+b_1) + b_2
\end{align}


On the decoder side, each layer performs two rounds of multi-head attention: the first one being self-attention over the representations of previously emitted tokens ($Q = K = V$), and the second being attention over the output of the final layer of the encoder ($Q$ are previous layer outputs, $K = V$ are $(h_1,...,h_N)$). The output of the final decoder layer for token $y_{l-1}$ is used to predict the following token $y_l$. Other components of the architecture such as sinusoidal positional encodings, residual connections and layer normalization are described in~\cite{vaswani2017attention}.





\vspace{-5pt}
\subsection{Convolutional layers}
Following previous work~\cite{dong2018speech, mohamed2019transformers, amodei2016deep}, we apply frequency-time 2-dimensional convolution blocks followed by max pooling to our audio features, prior to feeding them into the encoder, as seen in Fig. \ref{fig:block_diagram}. We can achieve significant savings in computation given that the resulting length of the input is considerably reduced and the computation required for self-attention layers scales quadratically with respect to the sequence length. 

Moreover, it has been shown that temporal convolutions are effective in modeling time dependencies~\cite{bai2018empirical}, and serves to encode ordering into learned high level representations of the input signal. Based on these observations, \cite{mohamed2019transformers} proposes to replace sinusoidal positional encodings in the Transformer with convolutions, employing 2D convolutions over spectrogram inputs and 1D causal convolutions over word embeddings in the decoder (pictured in Fig. \ref{fig:block_diagram}).

\vspace{-5pt}
\section{Model Compression}
\label{sec:quantization}

A simple approach to reducing computational requirements is to reduce the precision requirements for weights and activations in the model. It is shown in~\cite{polino2018model} that stochastic uniform quantization is an unbiased estimator of its input 
and quantizing the weights of a network is equivalent to adding Gaussian noise over parameters, which can induce a regularization effect and help avoid overfitting. Quantization has several advantages: 1) Computation is performed in fixed-point precision, which can be done more efficiently on hardware. 2) With 8-bit quantization, the model can be compressed up to 4 times of its original size. 3) In several architectures, memory access dominates power consumption, and moving 8-bit data is four times more efficient when compared to 32-bit floating point data. All three factors contribute to faster inference, with 2-3x times speed up~\cite{jacob2018quantization} and further improvements are possible with optimized low precision vector arithmetic. In this section, we summarize the quantization approach given in~\cite{prato2019fully}, which is deployed to fully quantize our simplified Transformer-based ASR.

\vspace{-5pt}
\subsection{Quantization scheme}
We use a uniform quantization function $Q: [a,b] \subseteq \Re \rightarrow [-2^{K-1},2^{K-1}-1] \subseteq \mathbb{Z}$ which maps real values (weights and activations) in the range of $[a,b]$ to $K$-bit signed integers: 
\begin{equation}
Q(x)= \operatorname{round}(\frac{x-a}{\Delta}) 
\label{quant}
\end{equation}
with $\Delta = \frac{b-a}{2^K-1}$.  In the case that $x$ is not in the range of $[a,b]$, we first apply the clamp operator:
\begin{align}
 \operatorname{clamp}(x;a,b) =\operatorname{min}(\operatorname{max}(x,a),b)
\end{align}
The de-quantization function $D(.)$ is given by:
\begin{align}
    D(x_Q)=  x_Q \times \Delta + a
    \label{de-quant}
\end{align}
where $x_Q = Q (x)$ refers to the quantized integer value corresponding to the real value $x$. 
During training, forward propagation simulates the effects of quantized inference by incorporating the de-quantized values of both weights and activations in the forward pass floating-point arithmetic operations. We then apply the quantization operation and the de-quantization operation according to eq.~\ref{quant} and eq.~\ref{de-quant} respectively to each layer. 
The clamping ranges are computed differently for weights and activations. For a weight matrix $X$, we set $a$ and $b$ to be $X_{min}$ and $X_{max}$ respectively. For activations, the clamping range depends on the $x$, the input to the layer. We calculate $[a,b]$ by keeping track of $x_{min}$ and $x_{max}$ for each mini-batch during training, and aggregating them using an exponential moving average with smoothing parameter set to $0.9$~\cite{prato2019fully}. 
Quantization of activations starts after a fixed number of steps (3000). This ensures that the network has reached a more stable stage and the estimated ranges do not exclude a significant fraction of values. We quantize to $K=8$-bit precision in our experiments.

\vspace{-5pt}
\subsection{Quantization choices}
The inputs and weights of all matrix multiplications are quantized while the addition operations are ignored since they do not lead to computational gains at inference time.
In the multi-head attention module, we quantize attention weights, softmax layer and scaled dot product 
and extend the same to all the layer norms in the model. 
For the position-wise feed forward network and convolution layers, we quantize the weights and activations.

\vspace{-5pt}
\section{Experiments}
\label{sec:experiments}

We use the open-source, sequence modelling toolkit \textit{fairseq} \cite{ott2019fairseq}. We conduct our experiments on LibriSpeech 960h~\cite{panayotov2015librispeech}, and follow the same setup as~\cite{mohamed2019transformers}: the input features are 80-dimensional log-Mel filterbanks extracted from 25ms windows every 10ms, and the output tokens come from a 5K subword vocabulary created with sentencepiece~\cite{kudo2018sentencepiece} ``unigram''. For fair comparison, we also optimize with AdaDelta~\cite{zeiler2012adadelta} with learning rate=1.0 and gradient clipping at 10.0, and run for 80 epochs, averaging checkpoints saved over the last 30 epochs. The dropout rate was set to 0.15. 

\vspace{-5pt}
\subsection{Comparison of Transformer variants}

We perform preliminary experiments comparing full-precision Transformer variants and choose one to quantize.  We start from \textbf{Conv-Context}~\cite{mohamed2019transformers} that proposes to replace sinusoidal positional encodings in the encoder and decoder with 2D convolutions (over audio features) and 1D convolutions (over previous token embeddings) respectively. Motivated by recent results in Transformer-based speech recognition~\cite{karita2019comparative} and language modelling~\cite{irie2019language}, we allocate our parameter budget towards depth over width, and retrain their model under the configuration of Transformer Base~\cite{vaswani2017attention}, namely: 6 encoder/decoder layers, $d_{model} = 512$, 8 heads, and $d_{ff} = 2048$. We obtain a satisfactory trade-off between model size and performance (Table \ref{table:convContext}), and adopt this configuration for the remainder of this work.

Next, we propose removing the 1D convolutional layers on the decoder side, based on previous work~\cite{irie2019language} demonstrating that the autoregressive Transformer training setup provides enough of a positional signal for Transformer decoders to reconstruct order in the deeper layers. We observe that removing these layers do not affect our performance, and reduce our parameter count from 52M to 51M. Finally, we add positional encodings on top of this configuration and see, counter-intuitively, that our performance degrades. These results are pictured in Table \ref{table:fullPrecision}.

\begin{table}[t]
   \caption{\ninept WER (\%) results of different hyperparameter configurations for Conv-Context. The first 2 rows are taken directly from~\cite{mohamed2019transformers}.}
   \label{table:convContext}
   \centering
   \resizebox{\columnwidth}{!}{
   \begin{tabular}{lrrrrrrrr}
   \toprule
   \multicolumn{1}{c}{\multirow{2}{*}{Model}} & \multicolumn{1}{c}{\multirow{2}{*}{$d_{model}$}} & \multicolumn{2}{c}{Layers} &  \multicolumn{1}{c}{\multirow{2}{*}{Params}} & \multicolumn{2}{c}{dev} & \multicolumn{2}{c}{test} \\
   \cmidrule(r){3-4} \cmidrule(r){6-7} \cmidrule(r){8-9} & & Enc & Dec  & & clean & other & clean & other \\
   \midrule
   \multirow{3}{*}{{\parbox{1.3cm}{Conv-Context}}} & \multirow{2}{*}{1024} & 16 & 6 & 315M & 4.8 & 12.7 & 4.7 & 12.9 \\
   & & 6 & 6 & 138M & 5.6 & 14.5 & 5.7 & 15.3 \\
   \cmidrule(r){2-9}
    & 512 & 6 & 6 & 52M & 5.3 & 14.9 & 5.7 & 14.8 \\
   \bottomrule
  \end{tabular}
  }
\end{table}

\begin{table}[t]
  \caption{\ninept Comparison of 3 full-precision model variants.}
  \label{table:fullPrecision}
  \centering
  \resizebox{\columnwidth}{!}{
  \begin{tabular}{lllcrrrr}
    \toprule
    \multicolumn{1}{c}{\multirow{2}{*}{Model}} &
    \multicolumn{1}{c}{\multirow{2}{*}{\parbox{0.6cm}{1D Conv}}} &
    \multicolumn{1}{c}{\multirow{2}{*}{\parbox{0.6cm}{Pos. enc.}}} &
    \multirow{2}{*}{Params} 
    & \multicolumn{2}{c}{dev}  & \multicolumn{2}{c}{test} \\
    \cmidrule(r){5-6} \cmidrule(r){7-8}
    & & & & clean &  other & clean & other \\
    \midrule
    Conv-Context & \cmark & \xmark & 52M & \textbf{5.3} & 14.9 & 5.7 & 14.8 \\
    \midrule
    Proposed & \xmark & \xmark & \multirow{2}{*}{51M} & 5.6 & \textbf{14.2} & \textbf{5.5} & 14.8 \\
    \hspace{.1cm}   + Pos. enc. & \xmark & \cmark & & 
    6.0 & 14.6 & 6.0 & \textbf{14.5} \\

    \bottomrule
  \end{tabular}
  }
\end{table}

 \vspace{-5pt}
\subsection{Quantization}
For quantization, we restrict our attention to our proposed simplified Transformer (no decoder-side convolutions or positional encodings), since it performs well and is the least complex of the Transformer variants. We compare the results of quantization-aware training to the full-precision model, as well as to the results of post-training quantization. In post-training quantization, we start from the averaged full-precision model, keep the weights fixed, and compute the clamping range $[a, b]$ for our activations over 1k training steps. To report the checkpoint-averaged result of quantization-aware training, we average the weights of quantization-aware training checkpoints, initialize our activation ranges $[a, b]$ with checkpoint averages, and adjust them over 1k training steps. In both cases, no additional updates are made to the weights.

Our results are summarised in Table \ref{table:quantized}. Our quantized models perform comparably to the full-precision model, and represent reasonable trade-offs in accuracy for model size and inference time. The last row of the table represents a result of 10x compression over the 138M parameter baseline with no loss in performance. Quantization-aware training scheme did not result in significant gains over post-quantization.

\begin{table}[ht!]
  \vspace{-7pt}
  \caption{\ninept Quantization results of our \textbf{proposed model} (no positional encodings or decoder-side convolutions).}
  \label{table:quantized}
  \centering
  \resizebox{\columnwidth}{!}{%
  \begin{tabular}{lcrrrrr}
    \toprule
    \multicolumn{1}{c}{\multirow{2}{*}{Model}} &
    \multicolumn{1}{c}{\multirow{2}{*}{\parbox{01.5cm}{Fully \phantom{aaa} quantized}}}
    &\multicolumn{2}{c}{dev}  & \multicolumn{2}{c}{test} \\
    \cmidrule(r){3-4} \cmidrule(r){5-6}
    &  & clean &  other & clean & other \\
    \midrule
    Full-precision & \xmark  &5.6 & \textbf{14.2} & 5.5 & \textbf{14.8} \\
    \midrule
    Post-training quant & \cmark  &5.6 & 14.6 & 5.6 & 15.1 \\
    Quant-aware training & \cmark  &\textbf{5.4} & 14.5 & 5.5 & 15.2 \\
    \bottomrule
  \end{tabular}
  }
  \vspace{-10pt}
\end{table}

\vspace{-5pt}
\section{Discussion}
\label{sec:discussion}

\subsection{Representing positional information}

The 3 Transformer variants explored in this work differ in how they present token-positional information to the decoder. We study their behaviour to get a better understanding as to why our proposed simplified model performs well.

\begin{figure}[t]
    \includegraphics[scale=0.150]{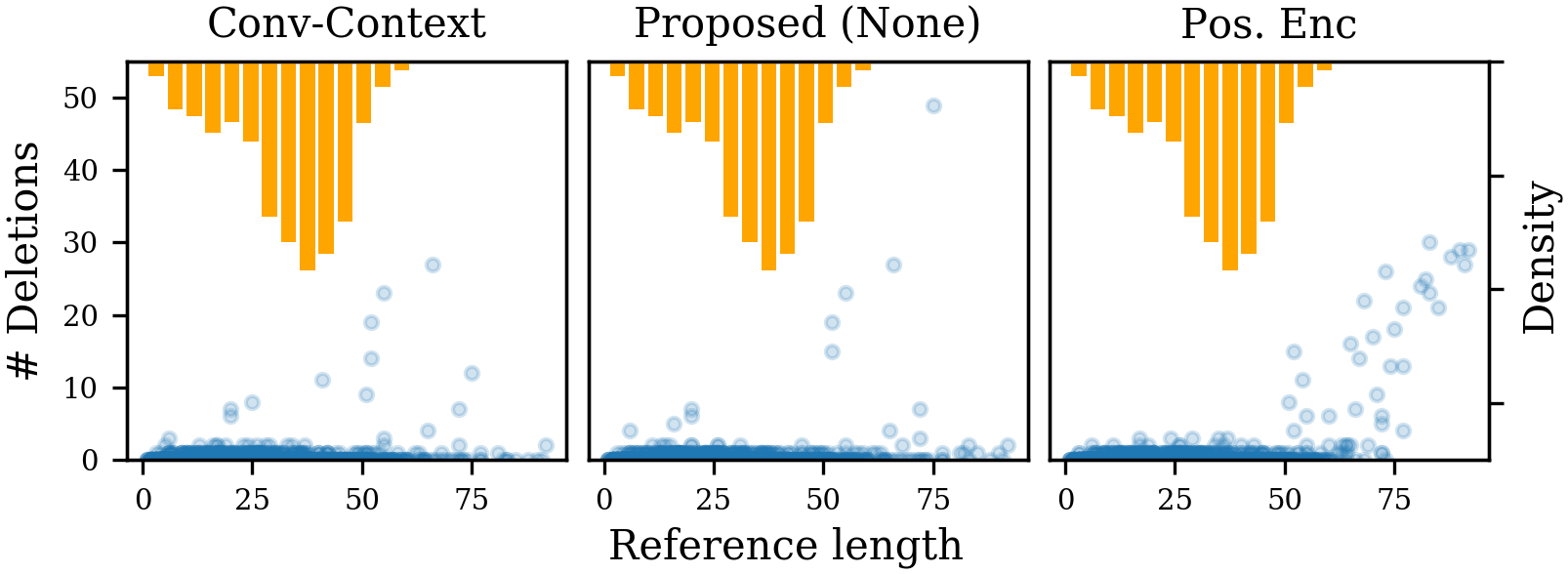}
    \vspace{-10pt}
    \caption{\ninept A plot of reference length vs. deletion for the dev-clean system output of our 3 models. The histograms in orange represent the length distribution of training transcriptions.}
    \label{fig:length_plot}
\end{figure}

We remark that \textit{sinusoidal position encodings hurt performance because of longer sequences at test time}. It has been observed that decoder-side positional encodings do worse than 1D convolutions \cite{mohamed2019transformers} (and also nothing at all, from our results). This performance drop is from under-generation; on dev-clean, our proposed model's WER increases $5.6 \rightarrow 6.0$ after adding positional encodings, with deletion rate increasing $0.7 \rightarrow 1.3$. Our plot in Fig. \ref{fig:length_plot} shows that this can be attributed to the inability of sinusoidal positional encodings to generalize to lengths longer than encountered in the training set.

Examining the same plot, we notice utterances with large deletion counts in the outputs of models without sinusoidal positional encoding. An example is shown in Fig. \ref{fig:skip_example}. \textit{Our models without sinusoidal positional encoding exhibit skipping.} We hypothesize the issue lies in the time-axis translation-invariance of decoder inputs: repeated n-grams confuse the decoder into losing its place in the input audio. Cross-attention visualizations between inputs to the final decoder layer and encoder outputs (left column of Fig. \ref{fig:attention}) support this hypothesis. We remark that being able to handle repetition is crucial for transcribing spontaneous speech. Imposing constraints on attention matrices or expanding relative positional information context are some possible approaches for addressing this problem.

Finally, we affirm the hypothesis proposed in \cite{irie2019language} that the Transformer with no positional encodings reconstructs ordering in deeper layers. The second column of Fig. \ref{fig:attention} show visualizations of cross-attention as we go up the decoder stack.

\begin{figure}[t]
     \centering
    \resizebox{\columnwidth}{!}{%
    \begin{tabular}{lp{80mm}}
    \toprule
    Reference & This second part is divided into two, for in the first \textbf{I speak of her as regards the nobleness of her} soul relating some of her virtues proceeding from her soul. In the second \textbf{I speak of her as regards the nobleness of her} body narrating some of her beauties here love saith concerning her. \\
    \midrule
    Conv\_Context & \textit{The} second part \textit{has} divided into two for in the first \textbf{I speak of her as regards the nobleness of her} \st{soul relating some of her virtues proceeding from her soul. In the second \textbf{I speak of her as regards the nobleness of her}} body narrating some of her beauties here love saith concerning her. \\
    \bottomrule
    \end{tabular}
    }
    \vspace{-10pt}
    \caption{\ninept An example of "skipping" taken from dev-clean. Punctuation is added for readability. In \textbf{bold} are repeated n-grams. \textit{The output of our proposed model is mostly identical to 1D Conv}. The model employing positional encodings makes no errors. }
    \label{fig:skip_example}
    \vspace{-13pt}
\end{figure}

\begin{figure}[t]
    \includegraphics[scale=0.145]{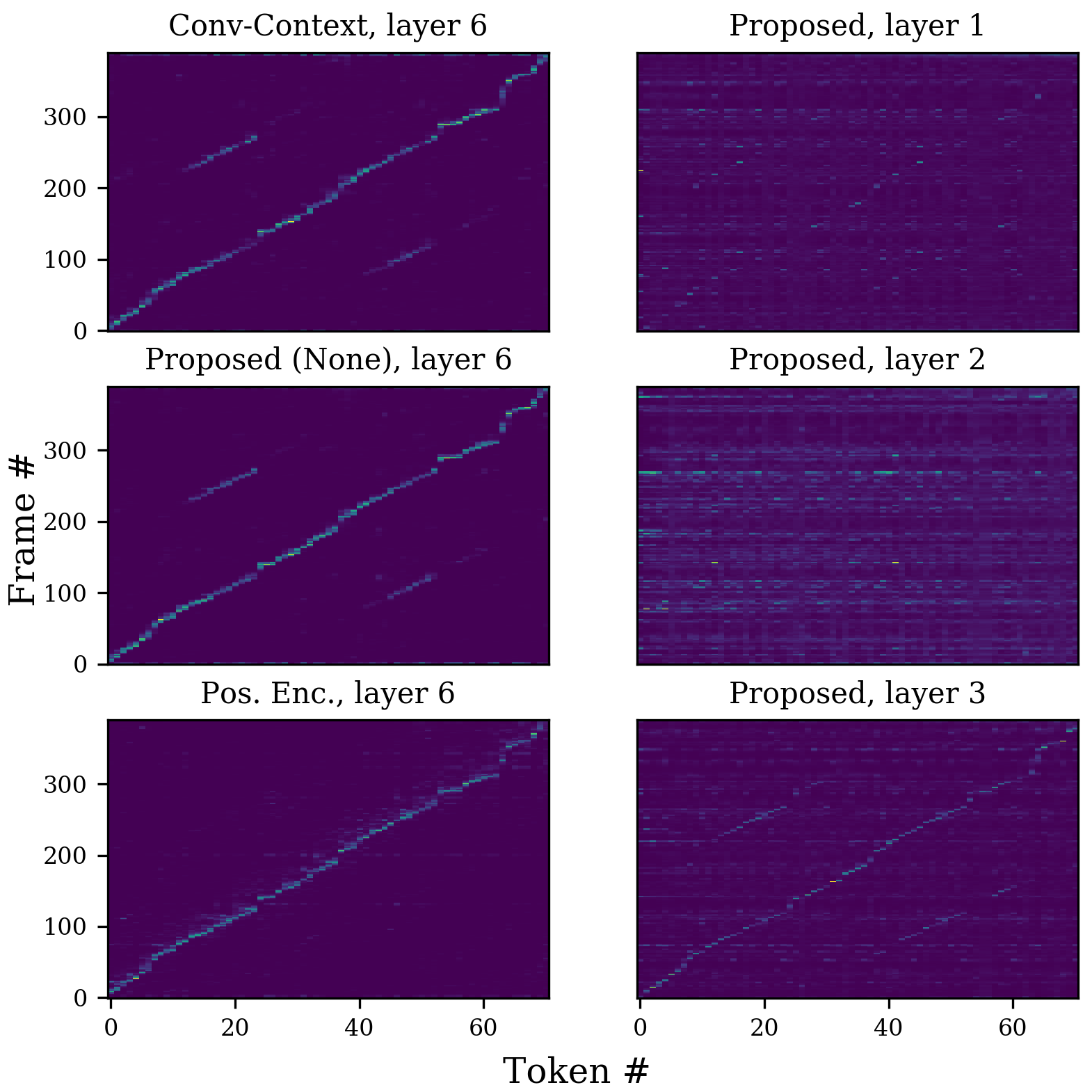}
    \vspace{-11pt}
    \caption{\ninept Decoder-encoder attention matrices for the utterance in Fig. \ref{fig:skip_example}. On the left column, we see the models without positional encoding sometimes exhibit bi-modality in attention distributions over the input audio. The transcription for the repeated section attends to both positions in the input audio. When decoding, the shorter path that skips the segment between the repetition has higher likelihood.}
    \label{fig:attention}
    \vspace{-13pt}
\end{figure}

\vspace{-5pt}
\subsection{Training the Transformer}

We observe no significant gain with quantization-aware training. Furthermore, it increases training time by more than 4x due to its expansion of our computational graph. We note that in post-quantization, the 1k steps used to fine-tune activation clamping ranges is very important. Without this step, system output is degenerate. \textit{In our experiments, we found that training with large batch sizes (80k audio frames) was necessary for convergence}. Similar optimization behaviour was observed across all experiments: a plateau at $\sim25\%$ frame-level accuracy followed by a jump to 80\% within 2 or 3 epochs. This jump was not observed when training with smaller batch sizes.


\vspace{-5pt}
\section{RELATION TO PRIOR WORK}
\label{sec:prior}

\textbf{Transformers for speech recognition.} Several studies have focused on adapting Transformer networks for end-to-end speech recognition. In particular,~\cite{dong2018speech,mohamed2019transformers} present models augmenting Transformers with convolutions.~\cite{karita2019comparative} focuses on refining the training process, and show that Transformer-based end-to-end ASR is highly competitive with state-of-the-art methods over 15 datasets. These studies focus only on performance, and do not consider trade-offs required for edge deployment. \\
\textbf{Compression with knowledge distillation.}~\cite{kim2019knowledge} proposes a knowledge distillation strategy applied to Transformer ASR to recover the performance of a larger model with fewer parameters. Distilled models still work in 32-bit floating point, and do not take advantage of faster, more energy-efficient hardware available when working with 8-bit fixed-point. Additionally, we believe this work is orthogonal to ours, and the two methods can be combined for further improvement. \\
\textbf{Transformer quantization.} Quantization strategies for the transformer have been proposed in the context of machine translation~{\cite{bhandare2019efficient, prato2019fully} and BERT \cite{zafrir2019q8bert}.
~\cite{prato2019fully} quantizes both weights and activations of the transformer for machine translation and language modeling tasks.
\textbf{Necessity of positional encodings.} For language modelling,~\cite{irie2019language} achieve better perplexity scores without positional encodings, and argue that the autoregressive setup used to train the Transformer decoder provides a sufficient positional signal.

\vspace{-8pt}
\section{Conclusion}
\label{sec:conclusion}
In this paper, we proposed a compact Transformer-based end-to-end ASR system, fully quantized to enable edge deployment. The proposed compact version has a smaller hidden size and no decoder side convolutions or positional encodings. We then fully quantize it to 8-bit fixed point. Compared to the 138M baseline we started from, we achieve more than 10x compression with no loss in performance. The final model also takes advantage of efficient hardware to enable fast inference. 
Our training strategy and model configurations are not highly tuned. Future work includes exploring additional training strategies and incorporating text data, as to bring highly performant, single-pass, end-to-end ASR to edge devices.

\vspace{-5pt}
\section{Acknowledgements}
We would like to thank our colleagues Ella Charlaix, Eyy\"ub Sari, and Gabriele Prato for their valuable insights and suggestions for quantization experiments.

\vfill

\bibliographystyle{IEEEbib}
\ninept
\bibliography{strings,refs}

\begin{thebibliography}{10}

\bibitem{graves2006connectionist}
Alex Graves, Santiago Fern{\'a}ndez, Faustino Gomez, and J{\"u}rgen
  Schmidhuber,
\newblock ``Connectionist temporal classification: labelling unsegmented
  sequence data with recurrent neural networks,''
\newblock in {\em Proceedings of the 23rd ICML}. ACM, 2006, pp. 369--376.

\bibitem{graves2014towards}
Alex Graves and Navdeep Jaitly,
\newblock ``Towards end-to-end speech recognition with recurrent neural
  networks,''
\newblock in {\em ICML}, 2014, pp. 1764--1772.

\bibitem{bahdanau2016end}
Dzmitry Bahdanau, Jan Chorowski, Dmitriy Serdyuk, Philemon Brakel, and Yoshua
  Bengio,
\newblock ``End-to-end attention-based large vocabulary speech recognition,''
\newblock in {\em 2016 IEEE ICASSP}. IEEE, 2016, pp. 4945--4949.

\bibitem{cho-etal-2014-learning}
Kyunghyun Cho, Bart van Merri{\"e}nboer, Caglar Gulcehre, Dzmitry Bahdanau,
  Fethi Bougares, Holger Schwenk, and Yoshua Bengio,
\newblock ``Learning phrase representations using {RNN} encoder{--}decoder for
  statistical machine translation,''
\newblock in {\em Proceedings of the 2014 EMNLP Conference)}, Doha, Qatar, Oct.
  2014, pp. 1724--1734, ACL.

\bibitem{sutskever2014sequence}
Ilya Sutskever, Oriol Vinyals, and Quoc~V Le,
\newblock ``Sequence to sequence learning with neural networks,''
\newblock in {\em Advances in neural information processing systems}, 2014, pp.
  3104--3112.

\bibitem{bahdanau2014neural}
D.~Bahdanau, K.~Cho, and Y.~Bengio,
\newblock ``Neural machine translation by jointly learning to align and
  translate,''
\newblock {\em arXiv preprint arXiv:1409.0473}, 2014.

\bibitem{chan2016listen}
W.~Chan, N.~Jaitly, Q.~Le, and O.~Vinyals,
\newblock ``Listen, attend and spell: A neural network for large vocabulary
  conversational speech recognition,''
\newblock in {\em 2016 IEEE ICASSP}. IEEE, 2016, pp. 4960--4964.

\bibitem{vaswani2017attention}
A.~Vaswani, N.~Shazeer, N.~Parmar, J.~Uszkoreit, L.~Jones, A.~N. Gomez,
  L.~Kaiser, and I.~Polosukhin,
\newblock ``Attention is all you need,''
\newblock in {\em Advances in neural information processing systems}, 2017, pp.
  5998--6008.

\bibitem{dong2018speech}
L.~Dong, S.~Xu, and B.~Xu,
\newblock ``Speech-transformer: a no-recurrence sequence-to-sequence model for
  speech recognition,''
\newblock in {\em 2018 IEEE ICASSP}. IEEE, 2018, pp. 5884--5888.

\bibitem{mohamed2019transformers}
A.~Mohamed, D.~Okhonko, and L.~Zettlemoyer,
\newblock ``Transformers with convolutional context for asr,''
\newblock {\em arXiv preprint arXiv:1904.11660}, 2019.

\bibitem{karita2019comparative}
S.~Karita, N.~Chen, T.~Hayashi, T.~Hori, H.~Inaguma, Z.~Jiang, M.~Someki,
  N.~E.~Y. Soplin, R.~Yamamoto, X.~Wang, et~al.,
\newblock ``A comparative study on transformer vs rnn in speech applications,''
\newblock {\em arXiv preprint arXiv:1909.06317}, 2019.

\bibitem{radford2019language}
A.~Radford, J.~Wu, R.~Child, D.~Luan, D.~Amodei, and I.~Sutskever,
\newblock ``Language models are unsupervised multitask learners,''
\newblock .

\bibitem{parmar2018image}
N.~Parmar, A.~Vaswani, J.~Uszkoreit, L.~Kaiser, N.~Shazeer, A.~Ku, and D.~Tran,
\newblock ``Image transformer,''
\newblock in {\em International Conference on Machine Learning}, 2018, pp.
  4052--4061.

\bibitem{jacob2018quantization}
B.~Jacob, S.~Kligys, B.~Chen, M.~Zhu, M.~Tang, A.~Howard, H.~Adam, and
  D.~Kalenichenko,
\newblock ``Quantization and training of neural networks for efficient
  integer-arithmetic-only inference,''
\newblock in {\em Proceedings of the IEEE Conference on CVPR}, 2018, pp.
  2704--2713.

\bibitem{lecun1990optimal}
Yann LeCun, John~S Denker, and Sara~A Solla,
\newblock ``Optimal brain damage,''
\newblock in {\em Advances in neural information processing systems}, 1990, pp.
  598--605.

\bibitem{han2015deep}
Song Han, Huizi Mao, and William~J Dally,
\newblock ``Deep compression: Compressing deep neural networks with pruning,
  trained quantization and huffman coding,''
\newblock {\em arXiv preprint arXiv:1510.00149}, 2015.

\bibitem{kim2019knowledge}
H.-G. Kim, H.~Na, H.~Lee, J.~Lee, T.~G. Kang, M.-J. Lee, and Y.~S. Choi,
\newblock ``Knowledge distillation using output errors for self-attention
  end-to-end models,''
\newblock in {\em ICASSP 2019-2019 IEEE International Conference on Acoustics,
  Speech and Signal Processing (ICASSP)}. IEEE, 2019, pp. 6181--6185.

\bibitem{kim2016sequence}
Yoon Kim and Alexander~M Rush,
\newblock ``Sequence-level knowledge distillation,''
\newblock {\em arXiv preprint arXiv:1606.07947}, 2016.

\bibitem{graves2012sequence}
Alex Graves,
\newblock ``Sequence transduction with recurrent neural networks,''
\newblock {\em arXiv preprint arXiv:1211.3711}, 2012.

\bibitem{He2018}
Y.~He, T.~N. Sainath, R.~Prabhavalkar, I.~Mcgraw, R.~Alvarez, D.~Zhao,
  D.~Rybach, Y.~Kannan, A.~Wu, and R~et~al. Pang,
\newblock ``Streaming end-to-end speech recognition for mobile devices.,''
\newblock 2018.

\bibitem{prato2019fully}
Gabriele Prato, Ella Charlaix, and Mehdi Rezagholizadeh,
\newblock ``Fully quantized transformer for improved translation,''
\newblock {\em arXiv preprint arXiv:1910.10485}, 2019.

\bibitem{panayotov2015librispeech}
V.~Panayotov, G.~Chen, D.~Povey, and S.~Khudanpur,
\newblock ``Librispeech: an asr corpus based on public domain audio books,''
\newblock in {\em 2015 IEEE International Conference on Acoustics, Speech and
  Signal Processing (ICASSP)}. IEEE, 2015, pp. 5206--5210.

\bibitem{sennrich2016neural}
Rico Sennrich, Barry Haddow, and Alexandra Birch,
\newblock ``Neural machine translation of rare words with subword units,''
\newblock in {\em Proceedings of ACL}, 2016, pp. 1715--1725.

\bibitem{amodei2016deep}
D.~Amodei, S.~Ananthanarayanan, R.~Anubhai, J.~Bai, E.~Battenberg, C.~Case,
  J.~Casper, B.~Catanzaro, Q.~Cheng, G.~Chen, et~al.,
\newblock ``Deep speech 2: End-to-end speech recognition in english and
  mandarin,''
\newblock in {\em International conference on machine learning}, 2016, pp.
  173--182.

\bibitem{bai2018empirical}
Shaojie Bai, J~Zico Kolter, and Vladlen Koltun,
\newblock ``An empirical evaluation of generic convolutional and recurrent
  networks for sequence modeling,''
\newblock {\em arXiv preprint arXiv:1803.01271}, 2018.

\bibitem{polino2018model}
A.~Polino, R.~Pascanu, and D.~Alistarh,
\newblock ``Model compression via distillation and quantization,''
\newblock {\em arXiv preprint arXiv:1802.05668}, 2018.

\bibitem{ott2019fairseq}
Myle Ott, Sergey Edunov, Alexei Baevski, Angela Fan, Sam Gross, Nathan Ng,
  David Grangier, and Michael Auli,
\newblock ``fairseq: A fast, extensible toolkit for sequence modeling,''
\newblock in {\em Proceedings of the 2019 Conference of the North American
  Chapter of the Association for Computational Linguistics (Demonstrations)},
  2019, pp. 48--53.

\bibitem{kudo2018sentencepiece}
T.~Kudo and J.~Richardson,
\newblock ``Sentencepiece: A simple and language independent subword tokenizer
  and detokenizer for neural text processing,''
\newblock {\em arXiv preprint arXiv:1808.06226}, 2018.

\bibitem{zeiler2012adadelta}
M.~D. Zeiler,
\newblock ``Adadelta: an adaptive learning rate method,''
\newblock {\em arXiv preprint arXiv:1212.5701}, 2012.

\bibitem{irie2019language}
K.~Irie, A.~Zeyer, R.~Schl{\"u}ter, and H.~Ney,
\newblock ``Language modeling with deep transformers,''
\newblock {\em arXiv preprint arXiv:1905.04226}, 2019.

\bibitem{bhandare2019efficient}
Aishwarya Bhandare, Vamsi Sripathi, Deepthi Karkada, Vivek Menon, Sun Choi,
  Kushal Datta, and Vikram Saletore,
\newblock ``Efficient 8-bit quantization of transformer neural machine language
  translation model,''
\newblock {\em arXiv preprint arXiv:1906.00532}, 2019.

\bibitem{zafrir2019q8bert}
Ofir Zafrir, Guy Boudoukh, Peter Izsak, and Moshe Wasserblat,
\newblock ``Q8bert: Quantized 8bit bert,''
\newblock {\em arXiv preprint arXiv:1910.06188}, 2019.

\end{thebibliography}

\end{document}